\documentclass[sigconf,natbib=true]{acmart}

\usepackage[ruled,linesnumbered]{algorithm2e}
\usepackage{multirow}
\usepackage{makecell}
\usepackage{enumitem}
\usepackage{caption}
\setitemize[1]{noitemsep,partopsep=0pt,parsep=0pt,topsep=0pt, leftmargin=10pt,
rightmargin=0pt}
\AtBeginDocument{%
  \providecommand\BibTeX{{%
    \normalfont B\kern-0.5em{\scshape i\kern-0.25em b}\kern-0.8em\TeX}}}

\setcopyright{acmcopyright}
\copyrightyear{2018}
\acmYear{2018}
\acmDOI{10.1145/1122445.1122456}

\acmConference[Woodstock '18]{Woodstock '18: ACM Symposium on Neural
  Gaze Detection}{June 03--05, 2018}{Woodstock, NY}
\acmBooktitle{Proceedings of the 32nd ACM International Conference on Information and Knowledge Management (CIKM),
  October 21--25, 2023, Woodstock, NY}
\acmPrice{15.00}
\acmISBN{978-1-4503-XXXX-X/18/06}

\begin{document}

\title{No Length Left Behind: Enhancing Knowledge Tracing for Modeling Sequences of Excessive or Insufficient Lengths}


\author{Moyu Zhang}
\affiliation{%
  \institution{Beijing University of Posts and Telecommunications}
  \city{Beijing}
  \state{Beijing}
  \country{China}
}
\email{zhangmoyu@bupt.cn}

\author{Xinning Zhu}
 \authornote{Corresponding Author}
\affiliation{%
  \institution{Beijing University of Posts and Telecommunications}
    \city{Beijing}
  \state{Beijing}
  \country{China}
}
\email{zhuxn@bupt.edu.cn}

\author{Chunhong Zhang}
\affiliation{%
  \institution{Beijing University of Posts and Telecommunications}
    \city{Beijing}
  \state{Beijing}
  \country{China}
}
\email{zhangch@bupt.edu.cn}

\author{Feng Pan}
\affiliation{%
  \institution{Beijing University of Posts and Telecommunications}
    \city{Beijing}
  \state{Beijing}
  \country{China}
}
\email{Pan_Feng@bupt.edu.cn}

\author{Wenchen Qian}
\affiliation{%
  \institution{Beijing University of Posts and Telecommunications}
    \city{Beijing}
  \state{Beijing}
  \country{China}
}
\email{wenchen@bupt.edu.cn}

\author{Hui Zhao}
\affiliation{%
  \institution{Beijing University of Posts and Telecommunications}
    \city{Beijing}
  \state{Beijing}
  \country{China}
}
\email{hzhao@bupt.edu.cn}

\begin{abstract}
Knowledge tracing (KT) aims to predict students' responses to practices based on their historical question-answering behaviors. However, most current KT methods focus on improving overall AUC, leaving ample room for optimization in modeling sequences of excessive or insufficient lengths. As sequences get longer, computational costs will increase exponentially. Therefore, KT methods usually truncate sequences to an acceptable length, which makes it difficult for models on online service systems to capture complete historical practice behaviors of students with too long sequences. Conversely, modeling students with short practice sequences using most KT methods may result in overfitting due to limited observation samples. To address the above limitations, we propose a model called Sequence-Flexible Knowledge Tracing (SFKT). Specifically, to flexibly handle long sequences, SFKT introduces a total-term encoder to effectively model complete historical practice behaviors of students at an affordable computational cost. Additionally, to improve the prediction accuracy of students with short practice sequences, we introduce a contrastive learning task and data augmentation schema to improve the generality of modeling short sequences by constructing more learning objectives. Extensive experimental results show that SFKT achieves significant improvements over multiple benchmarks, demonstrating the value of exploring the modeling of sequences of excessive or insufficient lengths. The code is available at https://github.com/zmy-9/SFKT.
\end{abstract}


\maketitle

\section{Introduction}
\begin{figure}[t]
  \centering
  \includegraphics[width=\linewidth]{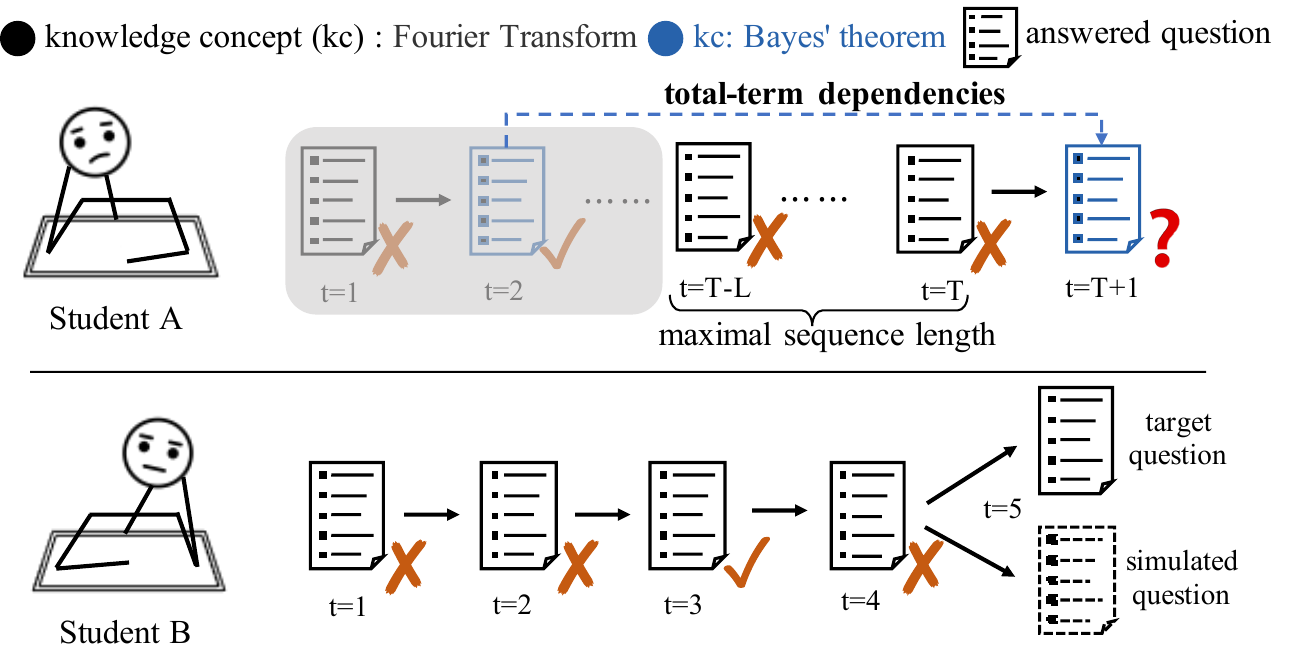}
  \caption{A toy example of KT. Assuming that the maximum length of input sequences for KT models is $L$, when student A's sequence length exceeds $L$, it becomes challenging to capture the impact of practice that occurred at time step t=2 on time step T+1. Conversely, student B's sequence length is short, KT models are more likely to overfit due to insufficient historical behaviors, and requires more training samples to alleviate the problem of data scarcity.}
  \label{example}
  \end{figure}

Knowledge tracing (KT) \cite{kt_task} has become a key component of online education systems due to the growing need for personalized education \cite{learn_rec1, learn_rec3}. As depicted in Figure \ref{example}, KT predicts the probability of a student correctly answering a question by analyzing their past practice sequences, allowing for monitoring of their knowledge state evolution \cite{kt_survey, learn_rec2,mob, bkt1}. Consequently, accurately modeling these practice sequences gathered by online education platforms is crucial for the KT task. While numerous methods have recently been proposed to enhance the ability to model practice sequences in the KT field, current KT methods primarily focus on improving overall AUC and leave ample optimization space for modeling sequences of excessive or insufficient lengths.

\begin{figure}[t]
  \centering
  \includegraphics[width=\linewidth]{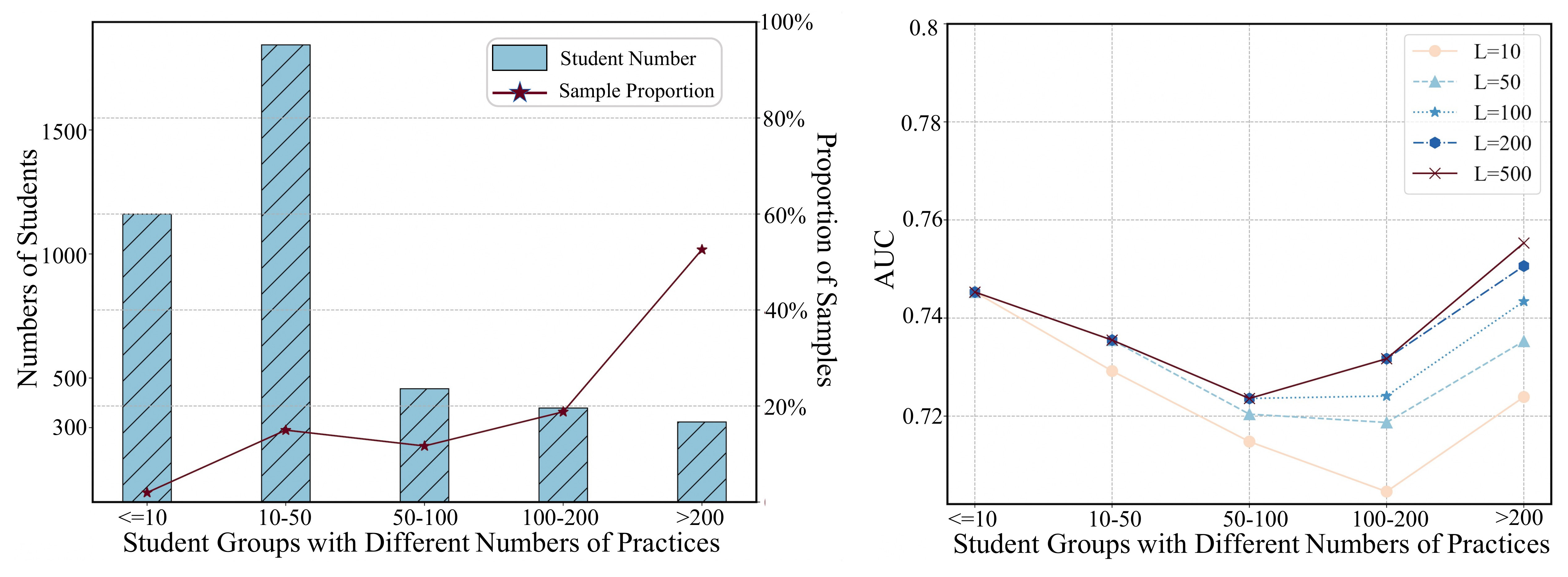}
  \caption{(a) Statistics of student practice data. (b) Predictive performance of DKT in the ASSIST2009\cite{assist09} dataset, where $L$ denotes the maximum input sequence length of DKT.}
  \label{statistic_example}
  \end{figure}

Recently, under the promotion of deep learning, KT models have predominantly adopted deep sequential structures \cite{kt_survey2, kt_survey}, such as LSTM \cite{lstm} or Transformer \cite{transformer}, to effectively capture long-term dependencies in student practice sequences. For instance, taking the representative KT method called Deep Knowledge Tracing (DKT) \cite{dkt} as an example, Figure \ref{statistic_example}(b) illustrates its prediction performance on student groups with different practice numbers under various maximum input sequence lengths. As seen from the results, longer maximum input sequence lengths improve DKT's prediction performance for students with long sequences (e.g., sequence longer than 200). However, increasing the maximum input sequence length leads to an exponential rise in computational costs. As a result, in real online education systems, KT models often have to limit the sequence length to hundreds, making it challenging for us to capture all historical practices for students with long practice sequences. Accordingly, it is crucial to explore efficient ways to capture all historical practice behaviors in long sequences without significantly increasing computational costs. For convenience, this paper uses the term \emph{total-term sequence} to refer to the entire historical practice sequence, while the term \emph{long-term sequence} is used to refer to the truncated sub-sequence within the maximum sequence length.

On the other hand, from Figure \ref{statistic_example}(b), we can find that increasing the maximum input sequence length does not enhance the model's predictive performance for students with short sequences (e.g., sequences shorter than 10). This is mainly due to the limited historical practice behaviors of short-sequence students, which makes it easy for the KT model to overfit. Furthermore, as shown in Figure \ref{statistic_example}(a), while short-sequence students constitute a significant proportion, the total number of their behaviors is limited. Thus, optimizing short-sequence modeling has little impact on overall AUC but can significantly enhance the online learning experience for these students. Consequently, addressing the challenge of modeling short sequences is crucial for the KT task.

To address the aforementioned challenges, we propose a novel model called \textbf{S}equence-\textbf{F}lexible \textbf{K}nowledge \textbf{T}racing (SFKT) to enhance the ability of KT models in modeling student sequences of excessive or insufficient lengths. SFKT incorporates a total-term encoder that captures all historical practice behaviors of students, maintaining constant time complexity regardless of sequence length. Additionally, SFKT introduces a contrastive learning task and data augmentation schema to jointly improve the model's generalization ability in short sequence modeling.

Specifically, SFKT utilizes two sequence encoders to monitor students' knowledge states over different time spans. The first encoder is named as \emph{Total-Term Encoder}, which models the student's knowledge state using prior statistical features based on the number of correct and incorrect practices related to knowledge concepts, similar to previous factor analysis methods \cite{ktm, mf_dakt}. However, previous factor analysis methods assumed a linear relationship between the number of practices and learning gains, which is inconsistent with common sense in education. Student learning usually undergoes quantitative changes leading to qualitative changes, and breakthrough progress is made when students practice to a certain quantity. Therefore, SFKT designs an auto-projector to learn the complex relationship between different practice times and learning gain, enhancing the representation ability of these prior statistical features. However, since this statistics-based approach may lose certain sequential information \cite{das3h}, SFKT introduces an encoder called \emph{Long-Term Encoder} to simulate the forgetting characteristics of students \cite{recent, forget}, where a position-aware attention mechanism is employed to capture the effect of the time interval between historical practice and target practice. Finally, the outputs of both encoders are jointly used to model the student's knowledge state and predict students' responses to practices.

However, the above sequence encoders cannot significantly enhance the model's ability to model short sequences due to the main challenge faced by short sequence modeling being the issue of insufficient samples. To this end, SFKT introduces a contrastive learning task and a data augmentation scheme to establish additional optimization objectives for the model, improving its generalization ability. Firstly, considering that although two sequence features output by the above two sequence encoders are modeled from different perspectives, the similarity between the two sequence features of the same student should be higher than the similarity between the sequence features of different students. Therefore, we propose to apply a contrastive loss to adjust the distance between the representation vectors of the outputs of sequence encoders to facilitate model learning. Secondly, we introduce a data augmentation scheme based on the assumption that if student B answered $q_5$ correctly, and $q_5$ is similar to $q_6$, then student B should have a high probability of answering $q_6$ correctly, as shown in Figure \ref{example}. Based on this assumption, we apply perturbation to each question in each sample to construct simulated samples, thereby mitigating the issue of insufficient samples when modeling short sequences and enhancing the model's generalization ability.

Finally, we conduct extensive experiments to compare our model with multiple state-of-the-art baselines on three public KT datasets. The experimental results strongly validate the superiority of SFKT.

The contributions of our paper can be summarized as follows:
\begin{itemize}
\item We point out that there is still optimization room in current KT field for modeling students' long or short sequences, and propose the SFKT to explore the problem of modeling sequences of excessive or insufficient lengths in a more targeted manner.
\item We propose a novel sequence encoder called Total-Term Encoder that can capture all of the student's historical practice behaviors in a flexible manner with a relatively low computational cost, regardless of the length of the student's practice sequence.
\item  We introduce a contrastive learning task and a data augmentation scheme to establish more optimization objectives, thereby improving the model's generalization ability in predicting students with too short practice sequences.
\end{itemize}  

\section{Related Work}

In recent decades, deep learning-based methods have become the mainstream direction in the field of KT \cite{ckt, akt, lpkt}. By designing effective deep network structures, these methods can better model student practice sequences than traditional KT methods. One of the most classical deep methods is Deep Knowledge Tracing (DKT) \cite{dkt} model, which is the first to apply long short-term memory network (LSTM) \cite{lstm} to the KT field and has been extended by the Dynamic Key-Value Memory model (DKVMN) \cite{dkvmn} to better capture long-term dependencies. Sequential Key-Value Memory Networks (SKVMN) \cite{skvmn} model combines the strengths of recurrent modeling capacity and memory capacity for modeling student learning. Later, inspired by the great success of the attention mechanism \cite{transformer}, many attention-based KT methods have emerged recently. Self-Attentive Knowledge Tracing (SAKT) \cite{sakt} is the first to apply self-attention mechanism to capture students' practices by calculating the relations between the target question and historical practices. To better model students' forgetting behaviors, Context-aware Attentive Knowledge Tracing (AKT) \cite{akt} designs a forgetting function based on the encoder structure of Transformer. HawkesKT \cite{hawkeskt} captures fine-grained temporal dynamics of different cross-skill impacts with the Hawkes Process \cite{Hawkes}. Later, Individual Estimation Knowledge Tracing (IEKT) \cite{iekt} model proposes the concept of students' cognition level and designs components to estimate the students' knowledge acquisition from practices. Similarly,  LFBKT \cite{lfbkt} comprehensively considers the factors that affect learning and forgetting behavior to model the knowledge acquisition, knowledge absorption and knowledge forgetting behaviors of students. Meanwhile, Difficulty Matching Knowledge Tracing (DIMKT) \cite{dimkt} model explores the question difficulty effect on learning by measuring students' feelings of the question difficulty to estimate their knowledge acquisition. Despite the superiority of these methods in modeling students' sequences, they overlook the increase in computational complexity with the increase of sequence length. At the same time, the above methods also tend to overfitting when modeling short sequences with insufficient samples.

Generally speaking, using statistical prior knowledge to assist model learning can enable modeling the complete sequence of students with lower computational costs. In the KT field, there is a traditional class of methods known as factor analysis methods \cite{irt}. These methods usually construct prior features by calculating the number of student practices involving various knowledge concepts, such as Additive Factor Model (AFM) \cite{afm1, afm2}, Performance Factor Analysis (PFA) \cite{pfa}, Knowledge Tracing Machine (KTM)\cite{ktm}, DAS3H \cite{das3h} and MF-DAKT \cite{mf_dakt}, to capture all of a student's historical practices with a constant-level time complexity. However, previous factor analysis methods consider the relationship between the number of student practices and the learning gain brought by practice as linear, which fails to model the complex learning process of students. Additionally, factor analysis methods have not solved the problem of insufficient data faced by short sequences modeling.

Thus, in this paper, we propose a method called Sequence-Flexible Knowledge Tracing (SFKT) to address the aforementioned issues. SFKT introduces a total-term encoder to learn the complex relationship between the number of practices and learning gains based on an auto-projector, thereby improving the model's ability to effectively model long sequences. Furthermore, SFKT introduces a contrastive learning task and a data augmentation scheme to enhance the generalization of the model in short sequence modeling. 

\section{Problem Formulation}
Given the historical practice sequence $ \chi_{u} = \left\{ p_1, ..., p_{t-1} \right\} $ of a student $u$, the KT task aims to predict the probability that $u$ answers question $q_t$ correctly at the next time step $t$, i.e. $ P \left ( y_{t} = 1 \mid u, q_{t}, \chi_{u} \right) $, where $p_{t-1} = \left (q_{t-1}, C_{q_{t-1}},  a_{t-1} \right )$. $ a_{t-1} \in \left\{ 0, 1 \right\} $ denotes the student's response to the question $q_{t-1}$ (1 means correct answer, and 0 means wrong answer). Since a question usually involves multiple knowledge concepts, each question has a set of pre-labeled knowledge concepts represented by $C_{q_{t-1}}$. 

The embedding vectors of $u$ and $q_t$ can be represented by $\boldsymbol{u} \in \mathbb{R}^{d_u}$ and $\boldsymbol{q}_{t} \in \mathbb{R}^{d_q}$, where $d_u$ and $d_q$ are dimensions of vectors. We use $\boldsymbol{C}_{q_{t-1}} \in \mathbb{R}^{d_c}$ to denote the averaged embedding vectors of knowledge concepts in $C_{q_{t-1}}$. Then, the embedding vector of the historical practice $p_{t-1}$ can be expressed as $\boldsymbol{p}_{t-1}$:
\begin{gather}
\boldsymbol{p}_{t-1} = \boldsymbol{W}_{0}(\boldsymbol{q}_{t-1} \oplus \boldsymbol{C}_{q_{t-1}} \oplus \boldsymbol{a}_{t-1}) + \boldsymbol{b}_{0}
\end{gather}
where $ \boldsymbol{a}_{t-1} \in \mathbb{R}^{d_a}$ denotes the embedding vector of $a_{t-1}$.$\oplus$ is the concatenation operation.  $\boldsymbol{W}_{0}$ and $\boldsymbol{b}_{0}$ are learnable parameters and $\boldsymbol{W}_{0} \in \mathbb{R}^{d \times (d_q + d_c+d_a)}$, $\boldsymbol{b}_{d} \in \mathbb{R}^{d}$. $d$ is the dimension of vector $\boldsymbol{p}_{t-1}$.
 
As previously mentioned, deep sequential models typically need to set a maximum input sequence length $L$. If the length of $ \chi_{u}$ is greater than $L$, we cannot input all the behaviors in $ \chi_{u}$ into the model when predicting $y_t$. Instead, we can only use $L$ adjacent behaviors as input, potentially leading to missing some historical practice information related to the target practice. Although increasing the maximum input sequence length can prevent this issue, it also results in an exponential increase in computational complexity, making deployment in real online systems almost impossible. This raises \textbf{\emph{Problem 1: how to flexibly model a student's practice sequence of excessive length}}. At the same time, if $t$ is small, the historical behaviors of $u$ are insufficient, making the model prone to overfitting. This leads to \textbf{\emph{Problem 2: 
how to enhance the KT method's generalization ability to model short sequences}}.

\begin{figure*}[t]
  \centering
  \includegraphics[width=\linewidth]{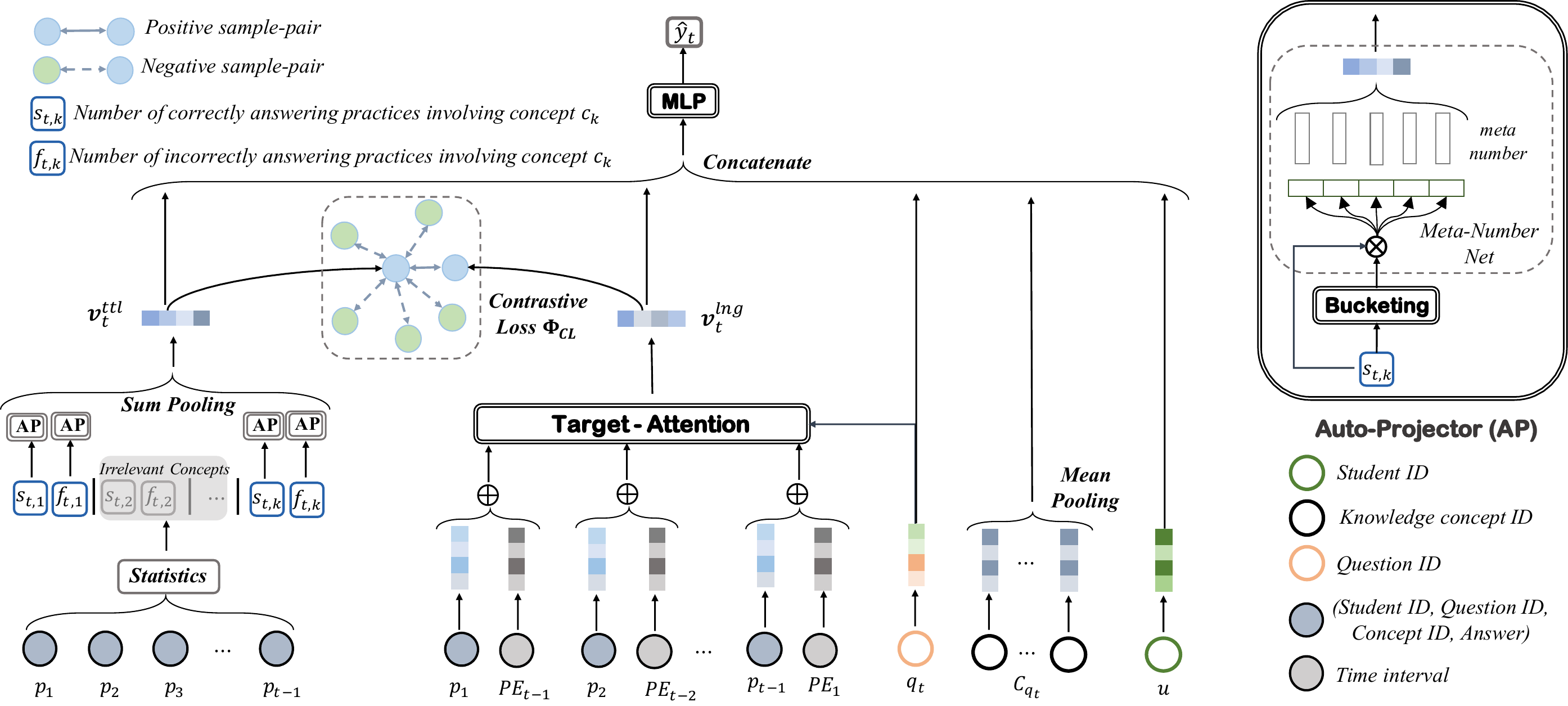}
  \caption{The main network architecture of the proposed SFKT model.}
  \label{kt:main_model}
\end{figure*}

\section{Method}
Previous KT methods did not sufficiently address situations where the student sequence was either too long or too short, leaving ample room for further optimization. To bridge this gap, we propose a model called Sequence-Flexible Knowledge Tracing (SFKT). Specifically, SFKT traces a student's knowledge state using two sequence encoders and predicts students' responses using the outputs of both sequence encoders. Additionally, SFKT jointly optimizes the model parameters based on a common prediction task, a contrastive learning task, and a data augmentation schema to improve the model's robustness and generalization for students with short sequences. The structure of SFKT is illustrated in Figure \ref{kt:main_model}.

\subsection{Trace Knowledge States}
To capture a student's complete practice history even when his historical practice sequence is too long, we propose a Total-Term Encoder that projects previous prior practice statistical features into the student's learning gains to trace students' knowledge states. However, the Total-Term Encoder may lose some sequential information of practices \cite{das3h}. Therefore, we also introduce a Long-Term Encoder to capture the sequential information of practices within time periods shorter than the maximum input sequence length.

\subsubsection{\textbf{Total-term Encoder}}
To consider a student's entire practice history, we conduct prior statistics based on their practice sequences and use an auto-projector to convert the prior statistical features into learning gains, thereby capturing the student's all historical behaviors with relatively low computational cost.

\emph{Prior Statistical Features}. Similar to previous factor analysis methods \cite{ktm, mf_dakt}, we record the number of times students answered correctly and incorrectly on each knowledge concept, denoted as $\boldsymbol{s}_{t} = (s_{t, 1},  s_{t, 2}, ..., s_{t, k}, ..., s_{t, N_c})$ and $\boldsymbol{f}_{t} = (f_{t, 1},  f_{t, 2}, ..., f_{t, k}, ..., f_{t, N_c})$, respectively, where $N_c$ is the total number of knowledge concepts. $s_{t, k}$ and $f_{t, k}$ represent the number of correct and incorrect answers to questions related to concept $c_k$ before time step t, respectively:
\begin{gather}
s_{t, k} = \sum_{i=1}^{t-1} [a_i \delta(c_k \in C_{q_i})] \\ 
f_{t, k} = \sum_{i=1}^{t-1} [(1-a_i) \delta(c_k \in C_{q_i})]
\end{gather}
where $\delta(c_k \in C_{q_i})=1$ when $c_k$ is in the concept set of $q_i$. 

\textbf{\emph{Auto-Projector}}. In previous factor analysis methods, the learning gains are represented by directly multiplying the number of practices by a vector representing the behavior of answering knowledge concepts \cite{ktm,mf_dakt}. This approach limits the model's representational power as it assumes a linear relationship between the number of practices and students' learning gains. In reality, the difference between answering correctly once and a hundred times is significant since quantitative changes lead to qualitative changes in learning. Recognizing that a simple linear relationship is insufficient to model the complex learning process, we design an auto-projector to map the intricate relationship between practice times and learning gains.

Assuming the input practice number is $x$, we use Min-Max Normalization to scale $x$ to a range of 0-1 and then employ a bucketing strategy to assign $x$ to one of B equal-width buckets, where values in the bucket share a representation vector, as follows:
\begin{gather}
\hat{x} =  \frac{log(x)-log(x^{min})}{log(x^{max})-log(x^{min})} \\ 
\boldsymbol{b}_{x} = Bucket (|\hat{x} (B-1)|),  \quad \boldsymbol{b}_{x} \in \mathbb{R}^{d}
\end{gather}
where $|\cdot|$ stands for the rounding down operation. $Bucket(\cdot)$ denotes that the integer $|\hat{x} (B-1)|$ is embed into corresponding bucket vector $\boldsymbol{b}_{x}$. However, this strategy may introduce noise. For instance, if we have a bucket for the number of practices between 1 and 10 and another one for values between 11 and 20, the number 11 is closer to 10, so we would expect that practicing 11 times should result in a learning gain more similar to practicing 10 times rather than 20 times. Nevertheless, according to the division strategy above, 11 and 20 will share the same bucket vector, which may lead to the model concluding that the learning gain from practicing 11 times is more similar to that from practicing 20 times.

To address the aforementioned issue, we first multiply the normalized value $\hat{x}$ by its corresponding bucket embedding vector $\boldsymbol{b}_{x}$, so that the representation vectors of numbers in the same bucket are slightly varied. Additionally, to further increase the difference between the representation vectors of numbers within the same bucket and enhance the similarity between adjacent numbers across different buckets, we employ a network called \emph{Meta-Number Net} for automatic clustering as follows:
\begin{gather}
\boldsymbol{\alpha}_{x} = \sigma(\boldsymbol{W}_{1}  (\hat{x} \boldsymbol{b}_{x}) + \boldsymbol{b}_{1})
\end{gather}
where $\sigma$ denotes the sigmoid function. $\boldsymbol{W}_{1}$ and $\boldsymbol{b}_{1}$ are learnable parameters and $\boldsymbol{W}_{1} \in \mathbb{R}^{M \times d}$, $\boldsymbol{b}_{1} \in \mathbb{R}^{M}$. The elements of $\boldsymbol{\alpha}_{x} \in \mathbb{R}^{M}$ indicate the probability that $\hat{x}$ belongs to M meta-numbers, where each meta-number can be seen as a cluster. Ultimately, we combine $\boldsymbol{\alpha}_{x}$ with the cluster embedding matrix $\boldsymbol{E}_{meta}  \in \mathbb{R}^{d \times M} $ to obtain the final representation vector of $\hat{x}$:
\begin{gather}
\boldsymbol{m}_{x} = \boldsymbol{E}_{meta} \boldsymbol{\alpha}_{x}, \quad \boldsymbol{m}_{x} \in  \mathbb{R}^{d}
\end{gather}
where $\boldsymbol{E}_{meta}$ is learnable. As the auto-projector is integrated into the KT framework, its parameters will be optimized in an end-to-end manner, with the final prediction loss aiding its ability to learn the relationship between practice numbers and learning gains.

Moreover, considering students' practices on different concepts should have different impacts on students' learning, we use matrices $\boldsymbol{E}_{s} \in \mathbb{R}^{N_c \times d}$ and $\boldsymbol{E}_{f} \in \mathbb{R}^{N_c \times d}$ to denote the behavior of correctly and incorrectly answering corresponding concepts, respectively. We combine the representation of answering a concept with the representation of the number of practices for that concept to capture a student's learning gain for the concept, as illustrated below:
\begin{gather}
\boldsymbol{v}_{t}^{ttl} =\boldsymbol{W}_{2} [ \sum_{c_k \in C_{q_t}}(\boldsymbol{m}_{s_{t, k}} \odot \boldsymbol{e}_{s, k}) \oplus \sum_{c_k \in C_{q_t}}(\boldsymbol{m}_{f_{t, k}} \odot \boldsymbol{e}_{f, k})] + \boldsymbol{b}_{2}
\end{gather}
where $\boldsymbol{v}_{t}^{ttl} \in \mathbb{R}^{d}$ denotes the extracted total-term feature for the target question $q_t$. $\odot$ denotes the element-wise product. $\boldsymbol{W}_{2}$ and $\boldsymbol{b}_{2}$ are learnable parameters and $\boldsymbol{W}_{2} \in \mathbb{R}^{d \times 2d}$, $\boldsymbol{b}_{2} \in \mathbb{R}^{d}$. $ \boldsymbol{e}_{s, k}$ and $ \boldsymbol{e}_{f, k}$ is the $k$-th row vector of $ \boldsymbol{E}_{s}$ and $ \boldsymbol{E}_{f}$, respectively.  $\boldsymbol{m}_{s_{t, k}}$ and $\boldsymbol{m}_{f_{t, k}}$ are outputs of auto-projector of correct and incorrect practice numbers $s_{t, k}$ and $f_{t, k}$ for $c_k$, respectively. In this way, the total-term encoder can capture a student's all practice behaviors with a time complexity of $O(x \times d)$, regardless of the length of the sequence, where $x$ denotes a constant. It is worth noting that there are two sets of auto-projector parameters used to model correct (e.g., $s_{t, k}$) and incorrect (e.g., $f_{t, k}$) practice times, respectively.

\subsubsection{Long-term Encoder}
Since the total-term encoder potentially loses sequential information \cite{recent}, we deploy a long-term encoder to emphasize the sequential information of historical practices over a period of time shorter than maximum sequence length, as follows:
\begin{gather}
\boldsymbol{Q}_{t} = \boldsymbol{W}_q (\boldsymbol{q}_{t} \oplus \boldsymbol{C}_{q_t}), \quad \boldsymbol{K}_{t} = \boldsymbol{W}_k (\boldsymbol{q}_{t} \oplus \boldsymbol{C}_{q_t})\\
\boldsymbol{v}_{t}^{lng} = \sum_{i=t-L}^{ t-1} \frac{exp((\boldsymbol{Q}_t)^{\top}  \boldsymbol{K}_{i})}{\sum_{{i}'=t-L}^{t-1} exp((\boldsymbol{Q}_t)^{\top}  \boldsymbol{K}_{{i}'})} (\boldsymbol{W}_v (\boldsymbol{p}_{i} + \boldsymbol{PE}_{t-i}))
\end{gather}
where $L$ denotes the maximum sequence length. $ \boldsymbol{PE}_{t-i}$ denotes the sine and cosine positional encoding \cite{transformer}, where $t-i$ denotes the time interval between current time step $t$ and previous time step $i$. $\boldsymbol{W}_q$, $\boldsymbol{W}_k$ and $\boldsymbol{W}_v$ denote the query, key and value matrix, respectively. $\boldsymbol{W}_{q} \in \mathbb{R}^{d \times d}$, $\boldsymbol{W}_{k} \in \mathbb{R}^{d \times d}$, $\boldsymbol{W}_{v} \in \mathbb{R}^{d \times d}$. 

\subsection{Model Prediction}
As $\boldsymbol{v}_{t}^{ttl}$ and $\boldsymbol{v}_{t}^{lng}$ have captured the student's knowledge state at time-step $t$, we further utilize both of them to predict the student's future performance on the target question $q_{t}$ at time-step $t$:
\begin{gather}
\hat{y}_{t} = \sigma(\boldsymbol{W}_{4} (\boldsymbol{W}_{3}(\boldsymbol{u} \oplus \boldsymbol{q}_{t} \oplus \boldsymbol{C}_{q_t} \oplus  \boldsymbol{v}_{t}^{ttl} \oplus \boldsymbol{v}_{t}^{lng}) + \boldsymbol{b}_{3}) + \boldsymbol{b}_{4})
\end{gather}
where $\hat{y}_{t}$ denotes the inferred probability of correctly answering $q_t$.  $\boldsymbol{W}_3$, $\boldsymbol{W}_4$, $\boldsymbol{b}_3$ and $\boldsymbol{b}_4$ are trainable parameters, and $\boldsymbol{W}_{3} \in \mathbb{R}^{2d \times 4d}$, $\boldsymbol{W}_{4} \in \mathbb{R}^{1 \times 2d}$,  $\boldsymbol{b}_{3} \in \mathbb{R}^{2d}$, $\boldsymbol{b}_{4} \in \mathbb{R}^{1}$. 

\subsection{Model Learning}
As previously mentioned, the main challenge when modeling short sequences is the scarcity of data, which leads to overfitting of the KT model. However, neither of the two sequence encoders mentioned above can address the issue of sparse samples, making it challenging to effectively enhance the KT method's ability to model short sequences. Drawing inspiration from recent breakthroughs in self-supervised learning \cite{sim-clr}, we propose the use of a contrastive learning task and a data augmentation scheme to alleviate the data sparsity issue and improve the model's generalization ability. 

\subsubsection{Prediction Loss}\label{pre_loss}
Since the primary objective of the KT task is to accurately predict the ability of a student correctly answering questions, we employ the cross-entropy loss function, which is a common loss function for the KT task, as the primary loss function:
\begin{equation}
\mathcal{L}_{Pred} = -\begin{matrix} \sum_{1 \leq t \leq T} y_t log\hat{y}_t + (1-y_t)log(1-\hat{y}_t) \end{matrix}
\end{equation} 
where $y_t$ is the true response label. 

\subsubsection{\textbf{Contrastive Loss}}
Although we introduce two sequence encoders to trace a student's knowledge status from different viewpoints, the sequence features generated by these encoders for the same student should show some level of similarity since they still indicate the same student's learning state. We anticipate that this similarity will be greater than the similarity between the sequence features of different students. In order to achieve this, we utilize a contrastive loss function \cite{cl1, cl2} to assist the model in learning the distance between the total-term and long-term sequence features.

Assuming the number of a batch of records is $\zeta$, we can obtain a batch of total-term features and long-term features as $\boldsymbol{V}^{ttl}= \left\{\boldsymbol{v}_1^{ttl}, \boldsymbol{v}_2^{ttl}, ..., \boldsymbol{v}_{\zeta}^{ttl}\right\}$ and $\boldsymbol{V}^{lng}= \left\{\boldsymbol{v}_1^{lng}, \boldsymbol{v}_2^{lng}, ..., \boldsymbol{v}_{\zeta}^{lng}\right\}$, respectively. The subscript here represents the index within the batch for clarity, which is different from the time-step in the sequence encoder. The contrastive loss function for the batch can be defined as:
\begin{gather}
\boldsymbol{h}_{\zeta}^{ttl} = ReLU (\boldsymbol{W}_{6} (ReLU (\boldsymbol{W}_{5}\boldsymbol{v}_{\zeta}^{ttl} + \boldsymbol{b}_{5})) + \boldsymbol{b}_{6}) \\
\boldsymbol{h}_{\zeta}^{lng} = ReLU (\boldsymbol{W}_{6} (ReLU (\boldsymbol{W}_{5}\boldsymbol{v}_{\zeta}^{lng} + \boldsymbol{b}_{5})) + \boldsymbol{b}_{6}) \\
\begin{aligned}
{\Phi}_{CL} &= \mathop{\mathbb{E}}\limits _{i \in |\zeta|} [-log \frac{e^{(\boldsymbol{h}_i^{ttl})^{\top}(\boldsymbol{h}_i^{lng}) /{\tau}}}{e^{(\boldsymbol{h}_i^{ttl})^{\top}(\boldsymbol{h}_i^{lng}) /{{\tau}}} + \sum \limits _{j \in |\zeta|} ^ {j \neq i} e^{(\boldsymbol{h}_i^{ttl})^{\top}(\boldsymbol{h}_j^{ttl})/{{\tau}}}}] \\
&+ \mathop{\mathbb{E}}\limits _{i \in |\zeta|} [-log \frac{e^{(\boldsymbol{h}_i^{ttl})^{\top}(\boldsymbol{h}_i^{lng})/{\tau}}}{e^{(\boldsymbol{h}_i^{ttl})^{\top}(\boldsymbol{h}_i^{lng})/{{\tau}}} + \sum \limits _{j \in |\zeta|} ^ {j \neq i} e^{(\boldsymbol{h}_i^{lng})^{\top}(\boldsymbol{h}_j^{lng})/{{\tau}}}}]
\end{aligned}
\end{gather}
$\boldsymbol{v}_{\zeta}^{ttl}$ and $\boldsymbol{v}_{\zeta}^{lng}$ will be projected to a representation space through a multi-layer network as $\boldsymbol{h}_{\zeta}^{ttl}$ and $\boldsymbol{h}_{\zeta}^{lng}$ to calculate the similarity, thereby enhancing the representation ability of the model \cite{sim-clr}. $\boldsymbol{W}_5$, $\boldsymbol{W}_6$, $\boldsymbol{b}_5$ and $\boldsymbol{b}_6$ are trainable parameters, and $\boldsymbol{W}_{5} \in \mathbb{R}^{2d \times d}$, $\boldsymbol{W}_{6} \in \mathbb{R}^{d \times 2d}$,  $\boldsymbol{b}_{5} \in \mathbb{R}^{2d}$, $\boldsymbol{b}_{6} \in \mathbb{R}^{d}$.  $\tau$ is the temperature coefficient used to control the discrimination of the model to negative sample pairs. 

\subsubsection{\textbf{Perturbation loss}}
Generally speaking, a student typically performs similarly on two similar questions. Thus, we can apply perturbations to the target question to generate simulated student responses to a simulated question similar to the original question. This schema enables us to generate new simulated samples for each record, which helps the model alleviate the overfitting issue caused by data scarcity. The process can be expressed as follows:
\begin{gather}
\boldsymbol{\tilde{q}}_{t} = \mathcal{D}(\boldsymbol{q}_{t}), \quad \boldsymbol{\tilde{C}}_{q_t} =\mathcal{D}(\boldsymbol{C}_{q_t}) \\
\boldsymbol{\tilde{v}}_{lng} = F_{lng}(\boldsymbol{\tilde{q}}_{t} \oplus \boldsymbol{\tilde{C}}_{q_t}, \left\{p_{t-L} , ..., p_{t-1}  \right \}) \\
\tilde{y}_{t} = \sigma(\boldsymbol{W}_{4} (\boldsymbol{W}_{3} (\boldsymbol{u} \oplus \boldsymbol{\tilde{q}}_{t} \oplus \boldsymbol{\tilde{C}}_{q_t} \oplus  \boldsymbol{v}_{t}^{ttl} \oplus \boldsymbol{\tilde{v}}_{t}^{lng}) + \boldsymbol{b}_{3}) + \boldsymbol{b}_{4})
\end{gather}
where the perturbation function $\mathcal{D}(\cdot)$ can take various forms, such as adding Gaussian Noise, randomly masking, and more. In this paper, we use the \emph{dropout} \cite{dropout} as the perturbation function, which has been proven to be an effective method for generating simulated samples in previous works \cite{sim-cse}. $F_{lng}(\cdot)$ denotes the long-term encoder, and $\tilde{y}_{t}$ denotes the prediction of the responses to the simulated question. Following the prediction loss, we utilize the cross-entropy loss to minimize the discrepancy between the prediction of the simulated question and the actual response to the original question: 
\begin{equation}
{\Phi}_{Pert}  = -\begin{matrix} \sum_{1 \leq t \leq T} y_t log\tilde{y}_t + (1-y_t)log(1-\tilde{y}_t) \end{matrix}
\end{equation} 

\subsubsection{Integration Objective Function} 
Building on the two regularization terms mentioned above, the overall objective function of SFKT, denoted as $\mathcal{L}$, can be expressed as:
\begin{gather}
\mathcal{L} = \mathcal{L}_{Pred} + \lambda_{CL}{\Phi}_{CL} + \lambda_{Pert} {\Phi}_{Pert} 
\end{gather}
where $\lambda_{CL}$ and $\lambda_{Pert}$ regulate the significance of the two regularization terms of contrastive loss and perturbation loss, respectively.

\begin{table}[t]
	\small
	\centering
	\begin{tabular}{cccc}
    		\hline & ASSIST2009 & ASSIST2012 & Algebra2005 \\
		\hline students & 3,883 & 28,325 & 567  \\
		questions & 17,737 & 53,079 &172,994 \\
		concepts & 123 & 265 & 111  \\
		records & 282,668 & 2,710,820 & 606,359   \\
		records /students & 66 & 96 & 1,069 \\
		\hline
	\end{tabular}
	\caption{Statistics of three benchmark datasets.}
	\label{kt:datasets}
\end{table} 

\begin{figure}[t]
  \centering
  \includegraphics[width=\linewidth]{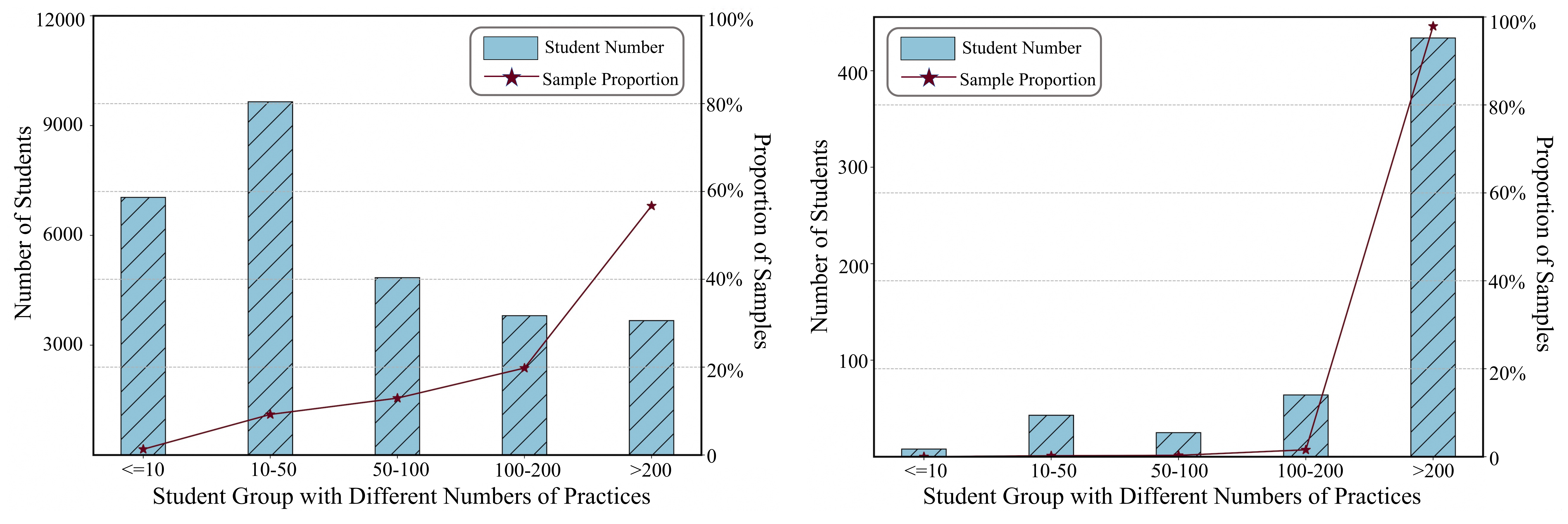}
  \caption{Statistics of ASSIST2012 and Algebra2005.}
  \label{kt:statistic}
\end{figure}

\section{Experiments}
This section presents several real-world datasets used to evaluate the effectiveness of our model, alongside multiple baseline models. Additionally, we provide a description of the training details and parameter settings used in the experiments. Next, we present the evaluation protocol and conduct a variety of experiments to answer the following research questions:\\ 
\textbf{Q1} Does SFKT indeed enhance the ability to model sequences of excessive or insufficient lengths? \\
\textbf{Q2} Does SFKT exhibit superiority compared to baselines that increase the length of the input sequence? \\
\textbf{Q3} What is the influence of each module in SFKT? \\
\textbf{Q4} What is the influence of each hyper-parameter in SFKT?

\subsection{Datasets} 
We evaluate the performance of KT methods on three real-world datasets: ASSIST2009, ASSIST2012, and Algebra2005. Our samples consist of question IDs, student IDs, knowledge concept IDs related to the corresponding questions, and students' responses. Table \ref{kt:datasets} lists the statistics of the datasets. Moreover, to justify the importance of enhancing the model's ability to handle sequences that are too long or too short, we perform statistical analysis on the three datasets, as shown in Figures \ref{statistic_example} and Figure \ref{kt:statistic}. These figures display the number of student groups with varying numbers of practice and the proportion of records for each group to the total dataset, respectively. Our analysis reveals that a significant proportion of students have short sequences (e.g., $\leq$ 10) in ASSIST2009 and ASSIST2012, while Algebra2005 has a large proportion of students with long sequences (e.g., $\geq$ 200). Based on the statistics, we conclude that a significant proportion of students have sequences that are too long or too short. Therefore, exploring methods to improve the model's ability to handle such sequences is necessary. A detailed description of the datasets is provided below: \\
 \textbf{$\bullet$ ASSIST2009\footnote{https://sites.google.com/site/assistmentsdata/home/2009-2010-assistment-data/skill-builder-data-2009-2010 \label{web}}}: This dataset is the most popular KT dataset gathered from the ASSISTments platform \cite{assist09}. It contains 282,668 samples with 109 concepts and 17, 737 questions from 3,883 students. Each question may involve one to four knowledge concepts. In ASSIST2009, a large proportion of students have short sequences (such as $\# \leq 10=26.3\%$) and long sequences  (such as $\# > 200=8.2$\%).  \\
\textbf{$\bullet$ ASSIST2012 \footnote{https://sites.google.com/site/assistmentsdata/home/2012-13-school-data-with}}.  This dataset also comes from the ASSISTments platform but has a larger number of students and observed samples. It is composed of 2,710,820 records, which involve 28,325 students. Each question is related to one knowledge concepts. ASSIST2012 also has a large proportion of students have short sequences (such as $\# \leq 10=22.4\%$) and long sequences  (such as $\# > 200=13.0\%$).\\
 \textbf{$\bullet$ Algebra2005}\footnote{https://pslcdatashop.web.cmu.edu/KDDCup/downloads.jsp}. This dataset stems from the KDD Cup 2010 EDM Challenge [36]. There are 567 students, 172,994 questions, and 111 concepts with 606,359 observed responses in total. Each question is related to one to seven knowledge concepts. In Algebra2005, sequences longer than 200 accounted for 75.6\%.
 
\subsection{Baselines}
\subsubsection{Factor Analysis Models}  Methods modeling students' learning by counting the students' practice times are listed in this part. \\
\textbf{$\bullet$ PFA} \cite{pfa} counts the number of students' right and wrong answers on concepts to model students' learning, respectively.\\
\textbf{$\bullet$ KTM} \cite{ktm} is the first to apply FM \cite{fm} into factor analysis models and becomes a generic framework.\\
\textbf{$\bullet$ DAS3H} \cite{das3h} introduces students' forgetting behavior by setting time windows to re-weight each practice.
\subsubsection{Deep Sequential Models} Methods extracting temporal fea- tures based on RNN-similar models are listed in this part. \\
\textbf{$\bullet$ DKT} \cite{dkt} is the first to apply the LSTM  to model the evolution of students' knowledge states.\\
\textbf{$\bullet$ DKVMN} \cite{dkvmn} extends DKT by using a key-value memory network to  update students' knowledge states.\\
\textbf{$\bullet$ SAKT} \cite{sakt} applies the encoder of Transformer to capture information of long-term students' practices. \\
\textbf{$\bullet$ AKT} \cite{akt} combines the self-attention mechanism with a monotonic attention mechanism to model sequences. \\
\textbf{$\bullet$ IEKT} \cite{iekt} sets cognition level and assesses students' knowledge acquisition to update model's input. \\
\textbf{$\bullet$ DIMKT} \cite{iekt} measures the question difficulty effect and improve KT performance by establishing the relationship between student knowledge state and question difficulty level.  \\
\textbf{$\bullet$ LFBKT} \cite{lfbkt} models the learning and forgetting behaviors based on educational psychology theory. 

\subsection{Experimental Settings} 
In our experiments, we set the embedding dimensions of students, questions, and concepts in SFKT to 64, along with a hidden vector dimension of $d=64$. We understand that the vector dimension settings could slightly impact the performance, but since our focus is on enhancing the model's ability to handle sequences, we will not be conducting separate experiments on this aspect. For the auto-projector, we set the number of buckets and meter numbers to 100, which can achieve relatively excellent performance on three datasets. We use the Adam \cite{adam} optimizer with a learning rate of 0.001 and set the mini-batch size to 24. The Xavier parameter initialization method \cite{initializer} is utilized to initialize the parameters of SFKT. Consistent with previous works \cite{dkt}, we set the maximum sequence length to 200 and truncate sequences longer than this into several sub-sequences based on the maximum length. 

Previous studies \cite{iekt, ktm} often exclude records of students who have practiced less than 10 times in their experiments as these records are deemed too short to model effectively. However, based on the statistical results above, we observe that many students practice less than 10 times, and thus we include all students' practice records in our experiments. For all datasets, we divided each student's behaviors in chronological order into a training set and a validation set, with the first 80\% of their behaviors used for training and validation, and the remaining 20\% used as the testing set.

\begin{table}[t]
	\small
	\centering
	\caption{Overall predictive performance of methods on three datasets.  * indicates p-value $\textless$ 0.05 in the significance test.}
	\begin{tabular}{c|cc|cc|cc}
		\hline \hline 
		\multirow{2}*{Method} & \multicolumn{2}{c|}{ASSIST2009} & \multicolumn{2}{c|}{ASSIST2012} & \multicolumn{2}{c}{Algebra2005}  \\
		 & ACC &AUC & ACC & AUC &ACC & AUC \\
		 \hline PFA&0.7104&0.7321&0.7181&0.7059&0.7866&0.7585 \\
		 KTM&0.7337&0.7609&0.7263&0.7642&0.8067&0.8013 \\
		 DAS3H&0.7345&0.7625&0.7281&0.7673&0.7976&0.7965 \\
		\hline DKT&0.7258&0.7446&0.7382&0.7356&0.8054&0.8091 \\
		 DKVMN&0.7302&0.7481&0.7359&0.7293&0.8015&0.8028\\	  
		SAKT&0.7310&0.7502&0.7388&0.7301&0.7987&0.7925 \\
		 AKT&0.7358&0.7650&0.7624&0.7793&0.8121&0.8209 \\
		IEKT &0.7328 & 0.7710 &0.7643 &0.7814 & 0.8072 & 0.8156\\
		DIMKT&0.7410&0.7783&0.7637&0.7885&0.8172&0.8283\\
		LFBKT&0.7397&0.7767&0.7624&0.7903&0.8153&0.8276\\
		\hline \hline SFKT&$\textbf{0.7580*}$&$\textbf{0.7867*}$&$\textbf{0.7680*}$&$\textbf{0.7973*}$ &$\textbf{0.8238*}$ &$\textbf{0.8397*}$   \\
		\hline \hline 
	\end{tabular}
	\vspace{-0.1cm}
	\label{kt:overall_performance}
\end{table}

\subsection{Evaluation Protocol}

\begin{table}[t]
	\small
	\centering
	\caption{The performance of KT methods on ASSIST2009.  * indicates p-value $\textless$ 0.05 in the significance test.}
	\resizebox{\linewidth}{!}{
	\begin{tabular}{c|ccccc}
		\hline \hline 
		Method & (0, 10] & (10, 50]  & (50, 100] & (100, 200]  & (200, $+\infty$) \\
		\hline PFA & 0.7137 & 0.7343 & 0.7123 &0.7033 &0.7329  \\
		KTM & 0.7432 & 0.7564 & 0.7484 &0.7476 &0.7611  \\
		DAS3H & 0.7443 & 0.7602 & 0.7538 &0.7468 &0.7617  \\
		\hline DKT&0.7453&0.7355&0.7236&0.7317&0.7506 \\
		 DKVMN&0.7343&0.7369&0.7239&0.7323&0.7515\\	  
		SAKT&0.7464&0.7319&0.7298&0.7238&0.7558 \\
		 AKT&0.7527&0.7530&0.7545&0.7513&0.7631\\
		IEKT &0.7512 & 0.7796 &0.7552 &0.7513 & 0.7681 \\
		DIMKT&0.7712&0.7898&0.7621&0.7687&0.7702\\
		LFBKT&0.7689&0.7912&0.7602&0.7623&0.7686\\
		\hline \hline SFKT&$\textbf{0.8203*}$&$\textbf{0.7933*}$&$\textbf{0.7682*}$&$\textbf{0.7728*}$ &$\textbf{0.7797*}$   \\
		\hline \hline 
	\end{tabular}
	}
	\vspace{-0.5cm}
	\label{kt:assist09_performance}
\end{table}  

Considering that one of the primary applications of KT is to rank questions based on the probability of students correctly answering them, we use Accuracy (ACC) and Area Under the ROC Curve (AUC) as the main metrics. However, as the focus of this paper is on modeling sequences of excessive or insufficient lengths, we recognize that the overall performance metrics may not fully capture the changes in the model's capability for these sequences. Thus, we aim to separately evaluate the prediction performance for students with too short and too long sequences. However, setting thresholds to divide sequences into "too short" and "too long" categories is challenging due to the lack of a clear standard. As a reference range, we define sequences with a length of less than 10 as "too short" sequences (which are often discarded in previous works), and sequences with a length greater than 200 as "too long" sequences (as previous works often set the maximum length to 200). To avoid controversy, we also report the models' prediction performance for the remaining ranges, including (10, 50], (50, 100], and (100, 200], to provide a more intuitive view of the experimental results. To ensure reliable experimental results, we ran each experiment five times with different random seeds and take the average value as the final result. Consistent with \cite{iekt}, we conduct a t-test \cite{t-test} under the ACC metric and a Mann-Whitney U test \cite{m-test} under AUC.

\begin{table}[t]
	\small
	\centering
	\caption{The performance of KT methods on ASSIST2012.  * indicates p-value $\textless$ 0.05 in the significance test.}
	\resizebox{\linewidth}{!}{
	\begin{tabular}{c|ccccc}
		\hline \hline 
		Method & (0, 10] & (10, 50]  & (50, 100] & (100, 200]  & (200, $+\infty$) \\
		 \hline PFA & 0.7133 & 0.7171 & 0.7094 &0.7164 &0.7075  \\
		KTM & 0.7409 & 0.7702 & 0.7653 &0.7684 &0.7593  \\
		 DAS3H & 0.7425& 0.7722 & 0.7665 & 0.7704 &0.7586  \\  
		\hline DKT&0.7331&0.7425&0.7358&0.7427&0.7338 \\
		 DKVMN&0.7243&0.7429&0.7289&0.7403&0.7315\\	  
		SAKT&0.7266&0.7404&0.7297&0.7357&0.7318 \\
		 AKT&0.7450&0.7868&0.7800&0.7832&0.7738\\
		IEKT &0.7428 & 0.7888 &0.7812 &0.7840 & 0.7762 \\
		DIMKT&0.7532&0.7953&0.7883&0.7925&0.7832\\
		LFBKT&0.7510&0.7929&0.7864&0.7881&0.7815\\
		\hline \hline SFKT&$\textbf{0.7926*}$&$\textbf{0.7981*}$&$\textbf{0.7892*}$&$\textbf{0.7957*}$ &$\textbf{0.7993*}$   \\
		\hline \hline 
	\end{tabular}
	}
	\label{kt:assist12_performance}
	\vspace{-0.1cm}
\end{table} 

\begin{figure}[t]
  \centering
  \includegraphics[width=0.7\linewidth]{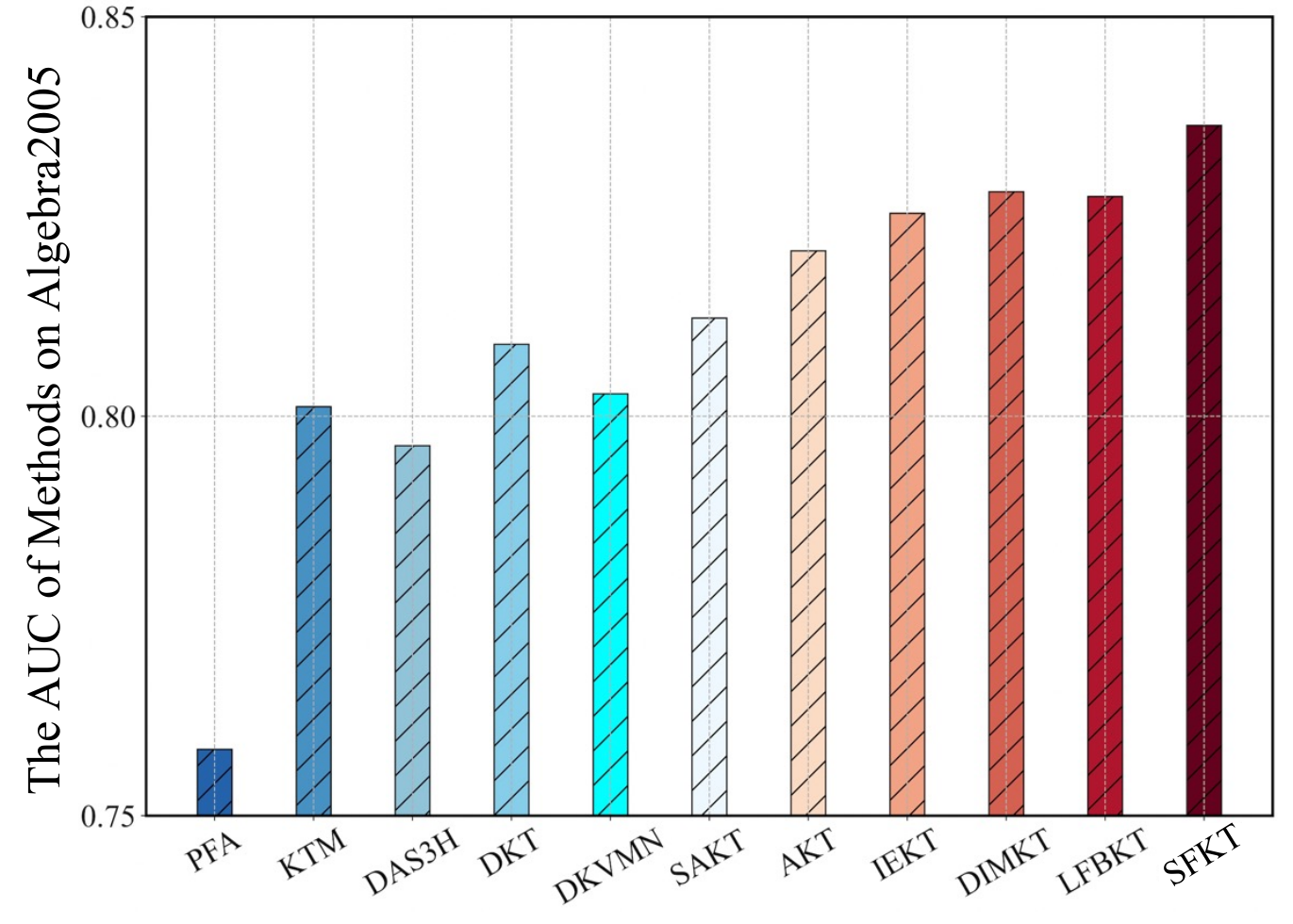}
  \caption{The performance of KT methods on Algebra2005.}
  \label{kt:algebra05_performance}
\end{figure}

\begin{figure*}[t]
  \centering
  \includegraphics[width=0.9\linewidth]{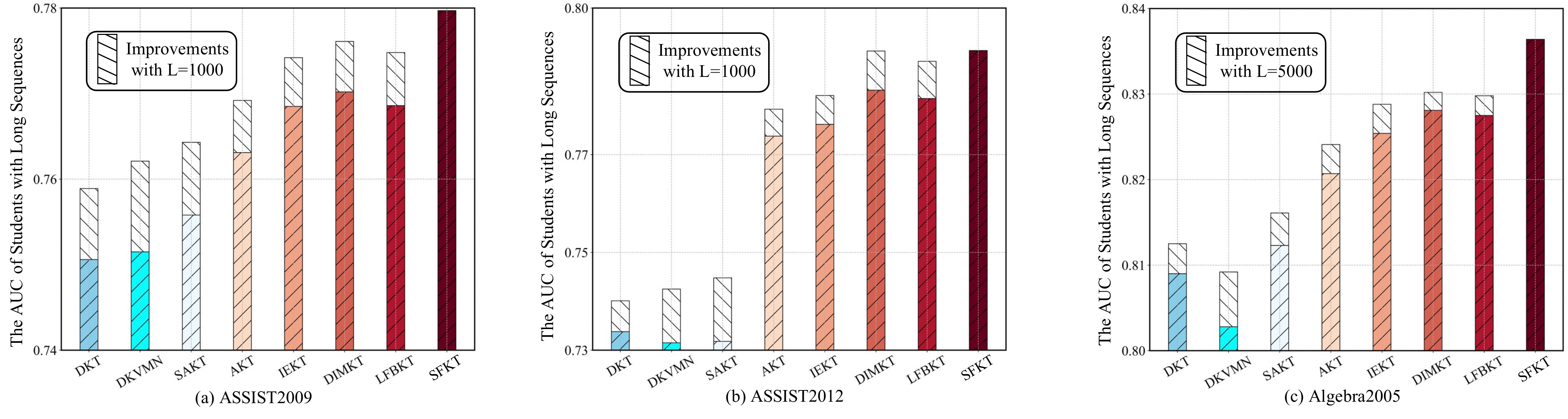}
  \caption{The impact of maximum input sequence length in modeling sequences longer than 200 on the three datasets.}
  \label{kt:length_auc}
  \vspace{-0.2cm}
\end{figure*}

\begin{table*}[t]
	\small
	\centering
	\caption{The performance of four variants on the three datasets.  We record the mean results over 5 runs. Std $ \leq$ 0.12\%. }
	\resizebox{\linewidth}{!}{
	\begin{tabular}{c|ccccc|ccccc|c}
		\hline \hline 
		\multirow{2}*{Method} & \multicolumn{5}{c|}{ASSIST2009} & \multicolumn{5}{c|}{ASSIST2012} &Algebra2005  \\
		& (0, 10] & (10, 50]  & (50, 100] & (100, 200]  & (200, $+\infty$) & (0, 10] & (10, 50]  & (50, 100] & (100, 200]  & (200, $+\infty$) & (200, $+\infty$) \\
		\hline SFKT&$\textbf{0.8203}$&$\textbf{0.7933}$&$\textbf{0.7682}$&$\textbf{0.7728}$ &$\textbf{0.7797}$ &$\textbf{0.7926}$&$\textbf{0.7981}$&$\textbf{0.7892}$&$\textbf{0.7957}$ &$\textbf{0.7993}$& $\textbf{0.8364}$  \\
		 w/o AP&0.8171&0.7882&0.7582&0.7653&0.7635&0.7863&0.7893&0.7805&0.7827 &0.7841 & 0.8172 \\
		w/o LTE&0.7824&0.7819&0.7573&0.7678&0.7742&0.7650&0.7890&0.7846&0.7877 &0.7912 & 0.8314  \\
		w/o CL&0.7887&0.7863&0.7622&0.7663&0.7732&0.7698&0.7958&0.7871&0.7910 &0.7933 & 0.8332 \\
		w/o Pert &0.7902 & 0.7832 &0.7628 &0.7672 & 0.7729&0.7763&0.7942&0.7851&0.7942 &0.7945 & 0.8351  \\
		\hline \hline 
	\end{tabular}
	}
	\label{kt:ablation}
	\vspace{-0.3cm}
\end{table*}  

\subsection{Performance on Predicting Responses (Q1)}
Table \ref{kt:overall_performance} displays the overall prediction performance of all methods on the three datasets, along with the statistical significance of our SFKT model against the best baseline model, with the highest results highlighted in bold. From the results in Table \ref{kt:overall_performance}, we can see that SFKT outperforms all baselines on both evaluation metrics for all datasets, indicating that modeling sequences that are either too long or too short can lead to global performance improvements.

To further analyze the model's performance in modeling sequences of different lengths, we group students based on the length of their sequences and made predictions for each group, as shown in Tables \ref{kt:assist09_performance}, \ref{kt:assist12_performance}, and Figure \ref{kt:algebra05_performance}. Since most students in the Algebra2005 dataset have sequences longer than 200, we only list the AUC of all methods for students with sequences longer than 200. Based on the experimental results of the three datasets, we find that our SFKT model outperforms all baseline models on students with sequence lengths shorter than 10 or longer than 200. This indicates that introducing the total-term encoder, the contrastive learning task, and the data augmentation schema has indeed enhanced the model's ability to model sequences that are too long or too short. Additionally, we find that SFKT's modeling ability for student groups with other sequence lengths compared to baseline models does not decrease, but even slightly improve, which is also reasonable. This is because the introduced modules in SFKT are equivalent to incorporating new practice information gain from different perspectives into existing models, which can assist in modeling sequences of any length.

In addition, we also observe an interesting phenomenon: factor analysis methods (KTM, DAS3H) perform better than DKT in modeling long sequences in the ASSIST2009 and ASSIST2012 datasets, but perform worse than DKT in the Algebra2005 dataset. The statistical results listed in Table \ref{kt:datasets} show that, on average, the length of practice sequences per student in Algebra2005 is longer than the other two datasets. While factor analysis methods can capture all these practices through statistical methods, they assume a linear relationship between practice number and learning gain, which may not accurately capture breakthrough learning gains from a large number of practices. This highlights the need to model the relationship between practice number and learning gain.

\subsection{Analysis on Input Sequence Length (Q2)}
We also conduct experiments to increase the maximum input sequence length of baselines and compare their performance to SFKT when not considering computational burden. To achieve this, we set the maximum sequence length of the ASSIST2009, ASSIST2012, and Algebra2005 datasets to 1000, 1000, and 5000, respectively, to ensure that almost all student sequences in the datasets could be fully input into the model. The results are presented in Figure \ref{kt:length_auc}, where the white box represents the improvement in performance compared to the input sequence length of 200. We observe that when the sequence length is increased, DIMKT and LFBKT could achieve similar or even better performance than SFKT. However, the time complexity of these methods became hundreds of times that of SFKT. We also find that increasing the input sequence length does not significantly improve the performance of the Algebra2005 dataset. This is because the average length of the practice sequence in Algebra2005 is too long, making it difficult for deep sequential models to capture the long-term dependencies between practices that are so far apart. Additionally, when the maximum sequence length was set to 200, the prediction performance of baseline models had already reached a high level, making further improvements more challenging on such a high basis.

\subsection{Ablation Study (Q3)}
To investigate the contribution of each module in SFKT, we conduct the ablation study on the datasets. We have four variants as follows: \\
\textbf{$\bullet$ w/o AP} removes the auto-projector module from the total-term encoder. That means, we directly multiple the practice numbers with representations of answering concepts and regard the relationship between practice number and learning gain as linear.  \\
\textbf{$\bullet$ w/o LTE} removes the long-term encoder from SFKT. That means, we trace students' knowledge states based on the total-term encoder and need to remove the contrastive task without LTE.  \\
\textbf{$\bullet$ w/o CL} removes the contrastive loss from the model learning.  \\
\textbf{$\bullet$ w/o Pert} removes the perturbation loss from the model learning. 

\begin{figure*}[t]
  \centering
  \includegraphics[width=0.9\linewidth]{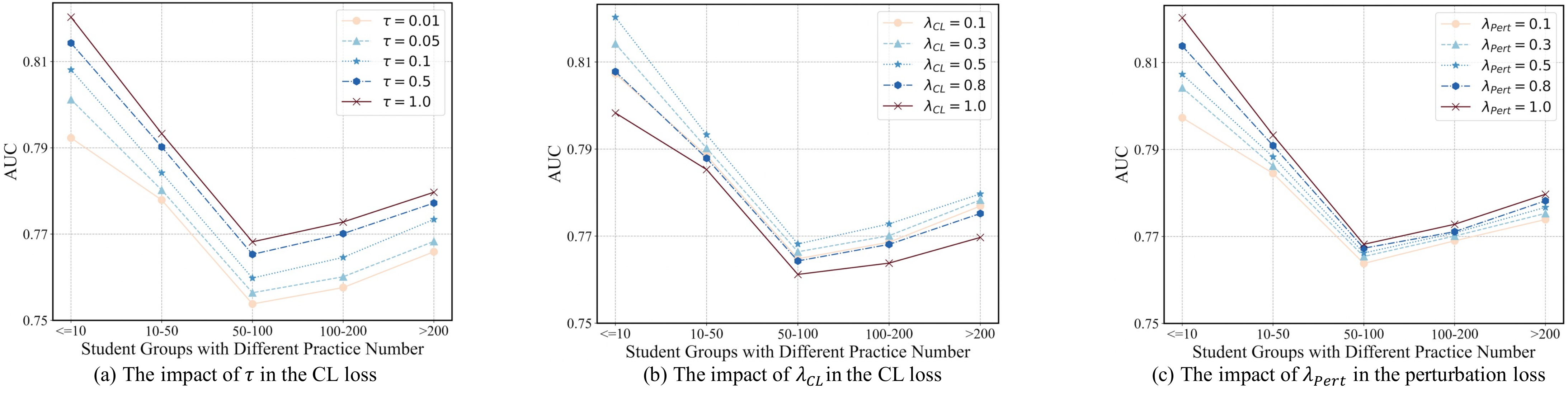}
  \caption{The performance of SFKT under different predefined hyper-parameters on the ASSIST2009 dataset.}
  \label{exp:hyper}
  \vspace{-0.1cm}
\end{figure*}

\begin{figure}[t]
  \centering
  \includegraphics[width=0.7\linewidth]{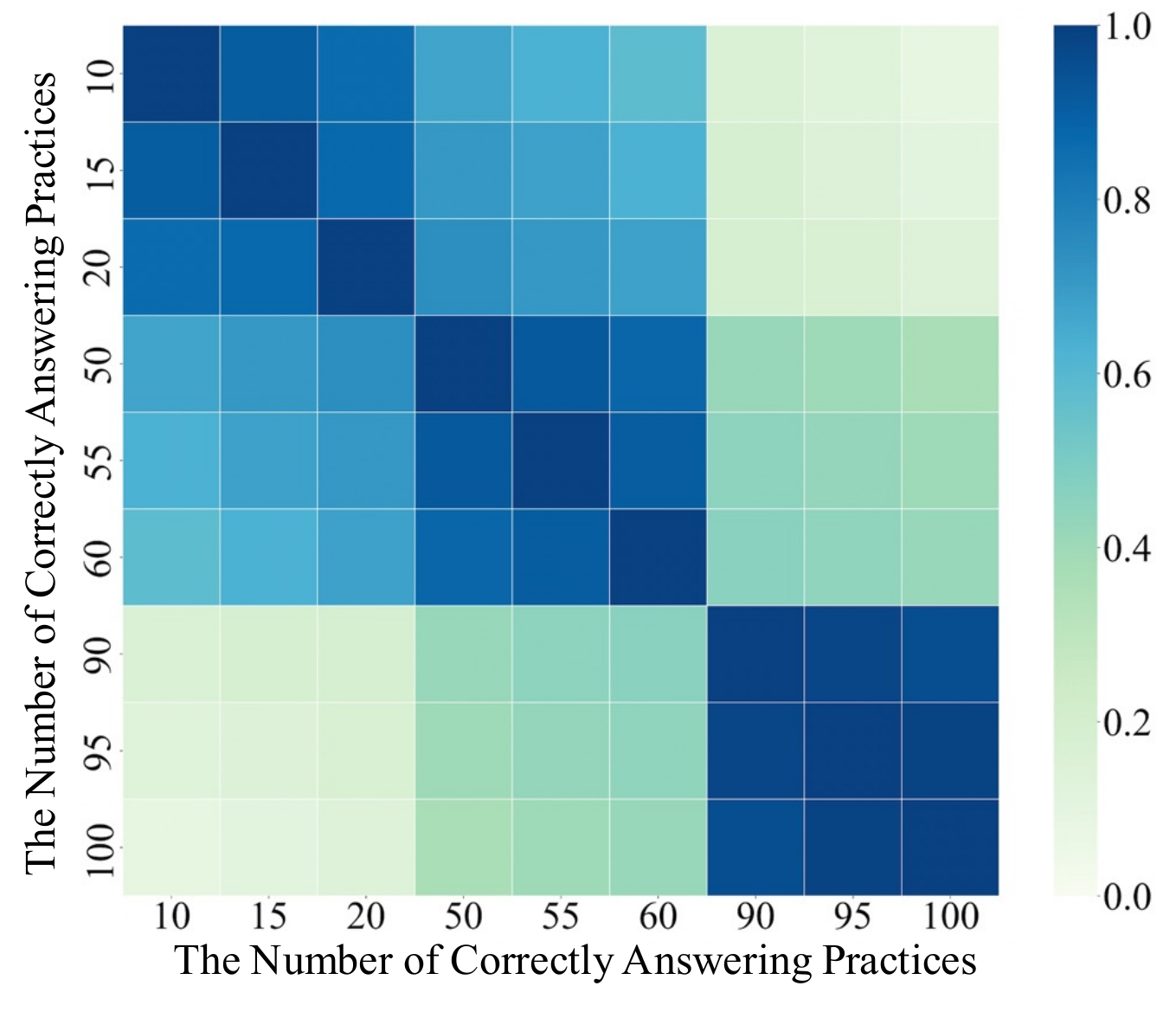}
  \caption{Similarity between different practice numbers.}
  \Description{It's the architecture of the MOKT model.}
  \label{exp:aup}
\end{figure}

Table \ref{kt:ablation} displays the performance of four SFKT variants. The results suggest that the performance of SFKT decreases regardless of which module is removed, implying that each module contributes to the predictive performance. Specifically, we find that the auto-projector (w/o AP) has the greatest impact on modeling sequences longer than 200, consistent with our hypothesis that SFKT's superiority in modeling long sequences is due to our attempt to learn the relationship between practice frequency and learning gain based on AP. Additionally, the Long-Term Encoder (w/o LTE) can improve the prediction performance of all sequence lengths. That's because that capturing nearby behaviors and practice intervals is crucial for modeling the students' learning progress \cite{akt, das3h}. Furthermore, we observe that the contrastive learning (w/o CL) and data augmentation (w/o Pert) tasks have a more significant impact on modeling short sequences. Our model's sufficient samples to learn to model long sequences lead to weak optimization effects of the two auxiliary losses in long sequence modeling. It is worth mentioning that all variant modules result in decreased performance when predicting students with sequence lengths between 10 and 200, indicating that these model structures do not compromise the modeling ability of other sequence lengths while enhancing the ability to model excessively long or short sequences. This further reinforces the rationality of our model structure.

\subsection{Hyper-Parameters Sensitivity Analysis (Q4)}
This section aims to investigate the sensitivity of hyperparameters, and all experiments are performed on the ASSIST2009 dataset. The results are illustrated in Figure \ref{exp:hyper}. Figure \ref{exp:hyper}(a) indicates that SFKT performs best when the temperature coefficient $\tau$ in the contrastive loss is set to 1.0. A smaller value causes SFKT to focus more on difficult sample-pairs. However, too small a value will constrain the similarity excessively, resulting in the loss of unique information of each sequence encoder, which contradicts the final prediction task. As KT aims to predict students' responses to practices, we set the coefficients of the two regularization terms in the objective function to be less than 1, enabling SFKT to focus on the prediction loss $\mathcal{L}_{Pred}$. We evaluate SFKT's performance on five values of $\lambda_{CL}$ and $\lambda_{Pert}$: 0.1, 0.3, 0.5, 0.8, and 1.0, as shown in Figures \ref{exp:hyper}(b) and \ref{exp:hyper}(c). SFKT's prediction performance was best when the coefficient $\lambda_{CL}$ of the contrastive learning task is 0.5. A smaller coefficient may not be sufficient to significantly affect model learning, while a larger coefficient will cause the optimization direction to deviate from the primary prediction task. Additionally, setting the coefficient of $\lambda_{Pert}$ in the perturbation task to 1.0 helps SFKT to perform best. We understand that because the perturbation we apply is slight, the perturbation task is similar to the primary prediction task, and assigning it the same weight can make it play a greater role.

\subsection{Visualization of Practice Numbers}
As mentioned earlier, we develop an auto-projector to learn the relationship between practice number and learning gain. To verify whether the auto-projector can solve this issue, we measure the similarity between the representation vectors of the projected practice numbers on the ASSIST2009 dataset. The lower the similarity between the representation vectors of two practice numbers with a large difference, the more the model can distinguish the learning gain brought by the two practice numbers. Figure \ref{exp:aup} illustrates the similarity between different practice number representations, where we observe that the closer the distance between the numbers, the higher the similarity between them. Practices with large differences, such as 10 and 100, have low similarity, indicating that the model distinguishes students practicing 10 times and practicing 100 times as significantly different, consistent with our design motivation. Additionally, the similarity between numbers within the range of 90 to 100 is higher than that within the range of 10 to 20, consistent with our common sense that continued practice after a student reaches a qualitative change slows down the change of the knowledge state gradually.

\section{Conclusions}
In this paper, we point out the optimizable space for modeling sequences of excessive or insufficient lengths in current KT field and propose a novel model named Sequential Flexible Knowledge Tracing (SFKT) to explore the optimization space for modeling sequences that are either too long or too short  in a more targeted manner. Firstly, we introduce a total-term encoder to capture a student's complete practice sequence with a constant level of time complexity, irrespective of the sequence length. Simultaneously, to compensate for the lack of sequential practice information, we combine the total-term encoder with a long-term encoder, which models sequential practice in a time period shorter than the maximum sequence length, to jointly trace students' knowledge states. Additionally, to enhance the model's capability to model short sequences, we introduce a contrastive learning task to adjust the distance between the outputs of the two sequence encoders and a data augmentation scheme to construct simulated samples to improve the model's generalization ability. Finally, we conduct extensive experiments to confirm the superiority of SFKT in modeling sequences of excessive or insufficient lengths.

\begin{acks}
This work is supported by 2022 Beijing Higher Education ``Undergraduate Teaching Reform and Innovation Project'' and 2022 Education and Teaching Reform Project of Beijing University of Posts and Telecommunications (2022JXYJ-F01).
\end{acks}

\end{document}